\documentclass[11pt]{article}

\usepackage[preprint]{acl}

\usepackage{cuted}
\usepackage{capt-of}
\usepackage{hyperref}
\usepackage{url}
\usepackage{graphicx}
\usepackage{adjustbox}
\usepackage{booktabs}
\usepackage{array}
\usepackage{booktabs, multirow, siunitx}
\usepackage{float} 
\usepackage[most]{tcolorbox}
\usepackage{amssymb} 
\usepackage{tabularray}
\usepackage{longtable}
\usepackage{xurl}      
\usepackage{hyperref}  
\usepackage{booktabs}
\usepackage{listings}
\usepackage{pifont}  
\usepackage{xcolor}  
\usepackage[dvipsnames]{xcolor}
\usepackage{supertabular}
\newcommand{\correct}{{\color{green}\ding{51}}} 
\newcommand{\wrong}{{\color{red}\ding{55}}}     

\usepackage{xcolor}
\usepackage[most]{tcolorbox}

\tcbset{
  promptbox/.style={
    enhanced,
    breakable,
    colback=black!2,
    colframe=black!35,
    boxrule=0.5pt,
    arc=2pt,
    left=2mm, right=2mm, top=1.2mm, bottom=1.2mm,
    coltitle=black,
    colbacktitle=black!8,
    fonttitle=\bfseries,
    attach boxed title to top left={xshift=1.5mm,yshift*=-1mm},
    boxed title style={sharp corners, boxrule=0pt},
    before skip=10pt,
    after skip=10pt,
  },
  promptinst/.style={
    colback=red!3,
    colframe=red!45,
    boxrule=0.4pt,
    arc=2pt,
    left=2mm, right=2mm, top=1mm, bottom=1mm,
  }
}

\newcommand{\PromptToyProblem}{%
Convert the point \((0,3)\) in rectangular coordinates to polar coordinates.
Enter your answer in the form \((r,\theta)\), where \(r > 0\) and \(0 \le \theta < 2\pi.\)

Present the answer in LaTeX format: \texttt{\textbackslash boxed\{Your answer\}}%
}

\usepackage{times}
\usepackage{latexsym}

\usepackage[T1]{fontenc}

\usepackage[utf8]{inputenc}

\usepackage{microtype}

\usepackage{inconsolata}

\usepackage{graphicx}

\title{Group Pattern Selection Optimization: Let LRMs Pick the Right Pattern for Reasoning}

\author{
Hanbin Wang$^{1}$\thanks{Equal contribution.},
Jingwei Song$^{2}$\footnotemark[1]\thanks{This work was done during an internship at Huawei Noah’s Ark Lab.},
Jinpeng Li$^{3}$\thanks{Corresponding author.},
Fei Mi$^{3}$,
Lifeng Shang$^{3}$ \\
$^1$Peking University \quad $^2$The University of Hong Kong \quad $^3$Huawei Noah’s Ark Lab\\
\small{
\href{mailto:wanghanbin95@stu.pku.edu.cn}{wanghanbin95@stu.pku.edu.cn},
\href{mailto:u3638265@connect.hku.hk}{songjingwei@connect.hku.hk}, 
\href{mailto:lijp.pku@gmail.com}{lijp.pku@gmail.com}
}
}

\begin{document}
\maketitle

\begin{abstract}
Large reasoning models (LRMs) exhibit diverse high-level reasoning patterns (e.g., direct solution, reflection-and-verification, and exploring multiple solutions), 
yet prevailing training recipes implicitly bias models toward a limited set of dominant patterns. Through a systematic analysis, we identify substantial accuracy variance across these patterns on mathematics and science benchmarks, revealing that a model’s default reasoning pattern is often sub-optimal for a given problem. To address this, we introduce Group Pattern Selection Optimization (GPSO), a reinforcement learning framework that extends GRPO by incorporating multi-pattern rollouts, verifier-guided optimal pattern selection per problem, and attention masking during optimization to prevent the leakage of explicit pattern suffixes into the learned policy.
By exploring a portfolio of diverse reasoning strategies and optimizing the policy on the most effective ones, GPSO enables the model to internalize the mapping from problem characteristics to optimal reasoning patterns. Extensive experiments demonstrate that GPSO delivers consistent and substantial performance gains across various model backbones and benchmarks, effectively mitigating pattern sub-optimality and fostering more robust, adaptable reasoning. All data and codes are available at \url{https://github.com/wanghanbinpanda/GPSO}.


\end{abstract}

\section{Introduction}
Recent advances in Large Language Models (LLMs), particularly those focused on complex reasoning, have yielded remarkable capabilities in solving challenging tasks across mathematics, science, and programming. Models like DeepSeek-R1 \citep{zhang2023deepseek} and OpenAI-o1 \citep{openai2024gpt} exemplify a new paradigm of "slow thinking," characterized by long, multi-step Chain-of-Thought (CoT) trajectories \citep{wei2022chain}.
A critical enabling factor behind this emergent behavior is the use of reinforcement learning (RL), with algorithms such as Proximal Policy Optimization (PPO) \citep{schulman2017proximalpolicyoptimizationalgorithms} and GRPO \citep{shao2024deepseekmathpushinglimitsmathematical} playing a central role. These training paradigms encourage models to explore, self-correct, and refine their reasoning on the fly, leading to impressive performance gains.

Inspired by these successes, a growing body of research has turned its attention to understanding the internal reasoning patterns adopted by these models. 
These reasoning patterns, or ``paradigms'', refer to the high-level, observable strategies a model employs to navigate a complex problem space, such as providing direct answers, decomposing problems, exploring alternative solutions, or employing tools. 
Several studies have systematically analyzed the cognitive behaviors of LLMs, revealing a rich spectrum of patterns such as self-reflection, backtracking, and exploration of multiple hypotheses \citep{wen2025thinkpatterns21ksystematicstudyimpact,gandhi2025cognitivebehaviorsenableselfimproving}.
Crucially, the reasoning patterns these models learn typically do not emerge spontaneously from scratch. Instead, they are shaped during the cold-start phase through human-designed prompts or explicitly reinforced by human preferences during reinforcement learning.
For example, \citet{chen2025mechanismreasoningpatternselection} analyze the evolution of these patterns before and after RL fine-tuning, finding that trained models tend to converge on a limited set of high-success-rate patterns. This observation leads us to a crucial, unaddressed question: \textbf{Are the reasoning patterns chosen by LRMs truly optimal for problem solving?}

\begin{figure*}[t]
    \centering
    \includegraphics[width=0.95 \linewidth]{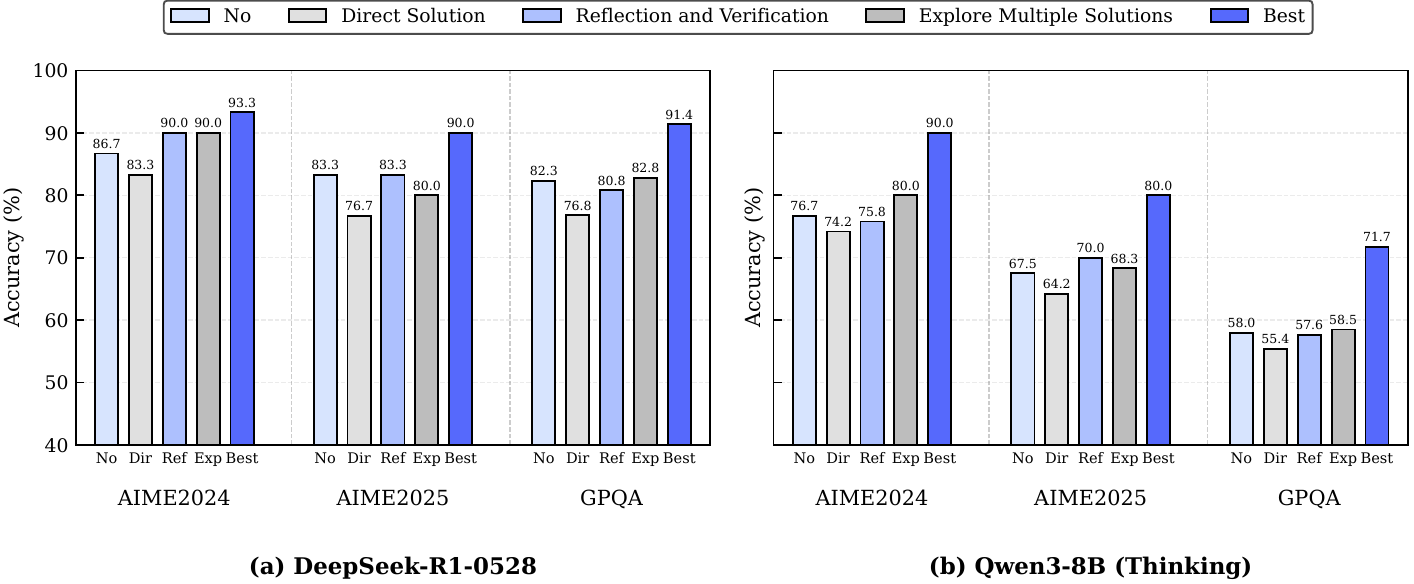}
    \caption{Comparison of model performance under different reasoning patterns on three benchmarks (AIME2024, AIME2025, and GPQA): (a) \textbf{DeepSeek-R1-0528} and (b) \textbf{Qwen3-8B (Thinking)}. \textbf{No}: No reasoning prompt, \textbf{Dir}: Direct solution, \textbf{Ref}: Reflection and verification, \textbf{Exp}: Explore Multiple solutions, \textbf{Best}: Pattern selected with the highest accuracy on each question.}
    \label{fig:best}
\end{figure*}

To investigate this, we conduct a comprehensive empirical study of reasoning trajectories. First, we perform a systematic analysis of the reasoning trajectories generated by seven state-of-the-art LLMs across mathematics, science, and code domains. Our analysis reveals that while LLMs possess the potential for diverse reasoning, they consistently default to a limited set of dominant patterns. Specifically, we find that the majority (approximately 98\%) of reasoning trajectories can be classified into three high-level categories: \textit{Direct Solution}, \textit{Reflection and Verification}, and \textit{Exploration of Multiple Solutions}. Interestingly, we observe that Reflection and Verification emerges as the default and primary reasoning pattern for most models, likely due to its robustness in self-correction (details can be found in Appendix \ref{app:distribution}).

Subsequently, we evaluate the performance of two high-performing LRMs (DeepSeek-R1-0528 and Qwen3-8B-Thinking) under these distinct reasoning patterns. They generate solutions using each of the three patterns through tailored, in-context prompts. The results, as illustrated in Figure \ref{fig:best}, reveal a striking finding: model performance varies significantly across different reasoning patterns. For instance, while Reflection and Verification might be optimal for some problems, Exploration of Multiple Solutions often yields substantially higher accuracy on tasks requiring novel insights. Critically, our results show that if LLMs were capable of dynamically selecting the most suitable pattern for each problem and outputting the best-performing trajectory, their overall performance could be enhanced by a substantial margin. This leads us to our core conclusion: \textbf{The reasoning patterns chosen by LRMs are not optimal.}

To bridge this gap, we propose Group Pattern Selection Optimization (GPSO), a novel training paradigm that teaches the model to intelligently select the optimal reasoning pattern for a given problem. Our method extends GRPO by incorporating multi-pattern exploration and optimal pattern optimization. During training, GPSO dynamically evaluates multiple candidate reasoning patterns for each problem. It then identifies the most effective pattern based on verifier-based signals and updates the policy model specifically on the rollouts of this optimal pattern. To ensure that the model learns the intrinsic mapping from problem to pattern—rather than overfitting to explicit pattern tokens—GPSO employs a gradient masking technique. This mechanism ensures that the explicit prompts used as exploration scaffolds do not leak into the learned policy, allowing the model to internally select the appropriate pattern on its own during reasoning. Through extensive experiments, we demonstrate that GPSO significantly outperforms existing methods and effectively addresses the sub-optimality issue in LLM reasoning.

Experimental results demonstrate that GPSO brings consistent and substantial improvements across diverse model backbones and reasoning benchmarks. For example, GPSO improves the average performance of Nemotron-Research-Reasoning-Qwen-1.5B from 55.4 to 58.0, a relative gain of +2.6\%. Similarly, DeepSeek-R1-Distill-Qwen-7B sees an increase from 55.6 to 58.7 (+3.1\%), while DeepSeek-R1-Distill-Llama-8B improves from 51.4 to 54.6 (+3.2\%). Moreover, our method proves highly effective even on the strongest baseline Qwen3-8B in our evaluation, which achieves an average improvement of 0.8\% after using GPSO. These findings establish GPSO as a model-agnostic and effective paradigm for maximizing reasoning potential.


\section{Related Work}
\subsection{Reinforcement Learning with Verifiable Rewards}
RLVR has emerged as a powerful and scalable post-training paradigm for large language models by leveraging rule-based or executable feedback, such as program execution results or logical consistency checks \citep{ouyang2022traininglanguagemodelsfollow,bai2022constitutionalaiharmlessnessai}. This approach bypasses the reliance on costly human-annotated reward models, showing strong improvements in reasoning-heavy domains like symbolic mathematics and code generation \citep{wang2025synthesizingsheetmusicproblems,chen2025symbolicgraphicsprogramminglarge}. The success of models like DeepSeek-R1 \citep{deepseekai2025deepseekr1incentivizingreasoningcapability}, which was trained with the GRPO algorithm \citep{shao2024deepseekmathpushinglimitsmathematical}, has inspired a surge of follow-up research \citep{he2025deltalnormalizationrethink,tang2025visualprogrammabilityguidecodeasthought,cheng2025k2thinkparameterefficientreasoning}. The researchers conduct in-depth studies on the design and robustness of the reward function in RLVR \citep{su2025crossingrewardbridgeexpanding,li2025implicitactorcriticcoupling,zhang2025tdrmsmoothrewardmodels}, the efficient utilization of data \citep{tang2025highdataefficiencyreinforcement,yang2025depthbreadthsynergyrlvrunlocking}, the balance mechanism between exploration and exploitation \citep{yang2025depthbreadthsynergyrlvrunlocking,wu2025invisibleleashrlvrescape,chen2025passktrainingadaptivelybalancing,wu2025invisibleleashrlvrescape}, and the cross-domain adaptation and multimodal reasoning \citep{chen2025enigmatascalinglogicalreasoning,xiao2025advancingmultimodalreasoningcapabilities,liang2025modomodomultidomaindatamixtures}.
\subsection{Sampling Strategies for Reinforcement Learning}
Efficient sample selection is critical for the convergence and performance of LLM fine-tuning, as it directly impacts which trajectories are prioritized for learning. Several prominent sampling strategies have been proposed.
Coarse-grained curriculum learning \citep{kimiteam2025kimik15scalingreinforcement,xie2025logicrlunleashingllmreasoning} gradually increases trajectory difficulty based on a competence-difficulty alignment score. LIMR \citep{li2025limrrlscaling} proposes Learning Impact Measurement (LIM) to prioritize problems whose expected learning progress best matches the current model trajectory. Prioritized Sampling \citep{kimiteam2025kimik15scalingreinforcement} weighs replay probability by TD-error or uncertainty, letting the agent reuse rare but informative transitions. Dynamic Sampling \citep{yu2025dapoopensourcellmreinforcement} monitors online pass rates and resamples low-variance trajectories until their outcomes are neither 0 nor 1, reducing redundancy at the cost of extra rollouts. MCTS-structured exploration \citep{make7030098} leverages tree search as a policy-improvement operator to steer deep RL toward high-value regions in vast action spaces, markedly boosting sample efficiency.

\subsection{Reasoning Patterns of Large Reasoning Models}

With the widespread adoption of RLVR, researchers have begun to investigate its effect on LLM behavior beyond simple performance metrics \citep{han2025selfalignedrewardeffectiveefficient,cheng2025k2thinkparameterefficientreasoning}. Some works begin to balance direct answers with extended thought processes to alleviate the problem of overthinking \citep{wu2025armadaptivereasoningmodel,fang2025thinklessllmlearnsthink,luo2025adar1hybridcotbileveladaptive,zhang2025adaptthinkreasoningmodelslearn}. However, few explore how reasoning patterns evolve during training. To address this, \citet{chen2025mechanismreasoningpatternselection} systematically investigates the role of RLVR for enhancing the reasoning capabilities of LLMs, discovering that their core advantage lies in optimizing the selection of existing high-success-rate reasoning patterns.
Building upon this crucial insight, our work is the first to propose a training framework that explicitly leverages and optimizes this pattern selection process to actively teach the model to pick the right pattern for each problem, thereby pushing the boundaries of LLM reasoning performance.
\section{Methodology}
\label{sec:methodology}
In this section, we introduce Group Pattern Selection Optimization (GPSO), which teaches the model to pick
the right pattern for reasoning. We first describe the preliminaries of Reinforcement Learning with Verifiable Rewards (RLVR) and then introduce our GPSO.

\begin{figure*}[t] \centering
    \includegraphics[width=0.9\textwidth]{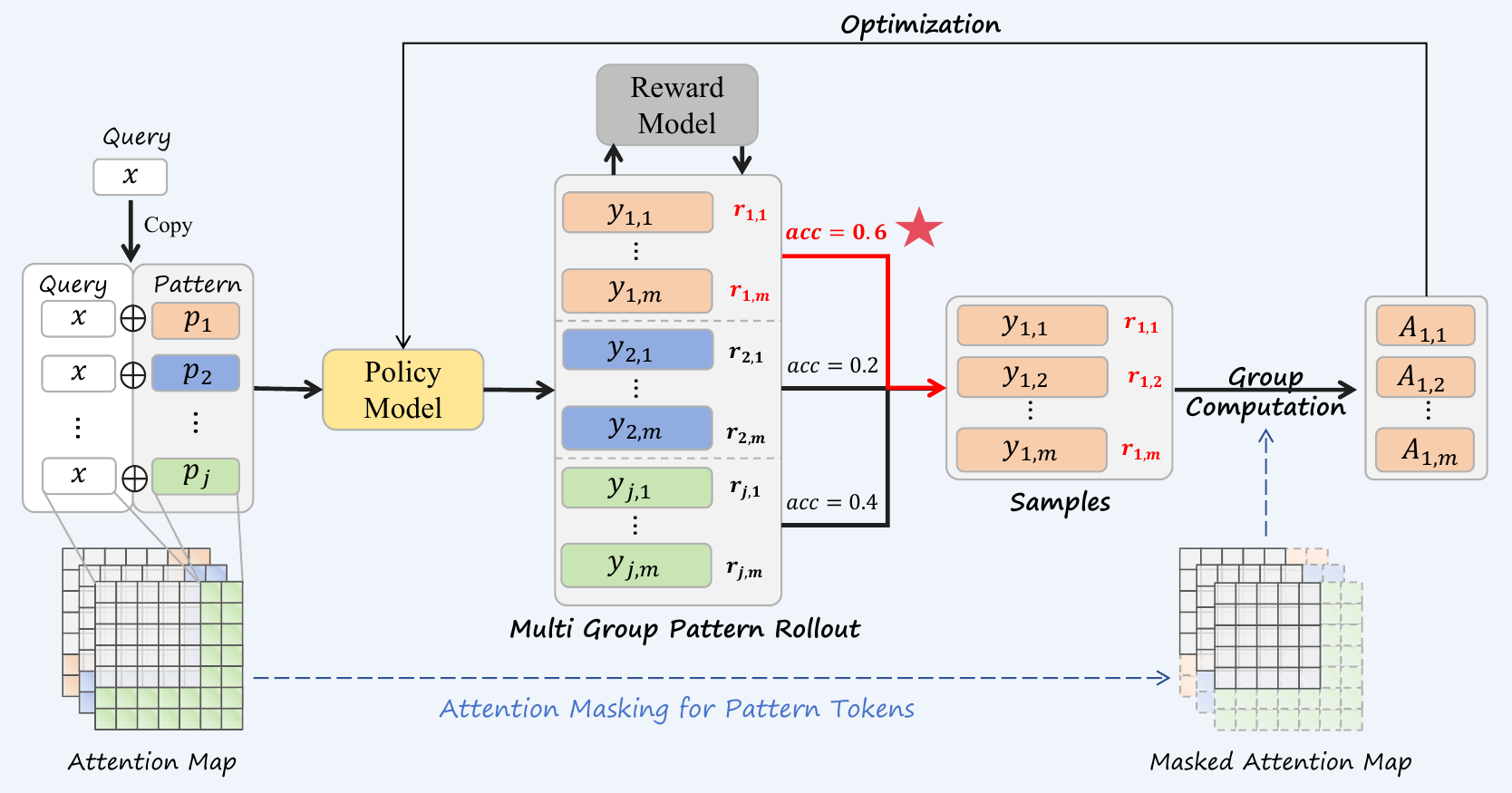}
    \caption{Overview of Group Pattern Selection Optimal (GPSO).} \label{fig:model}
\end{figure*}

\subsection{Preliminaries of RLVR}
Reinforcement Learning with Verifiable Rewards (RLVR) \citep{gao2024designingeffectiverlreward,lambert2025tulu3pushingfrontiers,deepseekai2025deepseekr1incentivizingreasoningcapability,kimiteam2025kimik15scalingreinforcement} refers to reinforcement learning optimization of models using rewards that can be automatically calculated using a rule-based verifier which assigns a scalar reward score to each generated response. Specifically, given a prompt $x$, the policy $\pi_\theta$ generates a reasoning trace $z$ followed by a final answer $y$. A verifier computes a reward $r = Verifier(y,y^*)$. Training proceeds via standard RL algorithms (e.g., PPO \citep{schulman2017proximalpolicyoptimizationalgorithms} or GRPO \citep{shao2024deepseekmathpushinglimitsmathematical}) to maximize the expected verifier reward, i.e.:
\begin{equation}
\max_\theta \; \mathbb{E}_{z, y \sim \pi_\theta(\cdot|x)} \big[ Verifier(y, y^*) \big]
\end{equation}
where $Verifier$ is a rule-based function that compares the model output $y$ against the reference answer $y^*$ and returns a scalar score.


In this paper, we adapt Group Relative Policy Optimization (GRPO) as our reinforcement learning objective. GRPO is a PPO-like actor-only algorithm that omits the learning of a separate value function. For each prompt \(x\), it samples a group of \(G\) reasoning traces and answers \(\{(z_i, y_i)\}_{i=1}^G\), each yielding a scalar reward \(r_i = \text{Verifier}(y_i, y^*)\). The optimization objective is:




\begin{equation}
\begin{split}
    \mathcal{L}_{\text{GRPO}}(\theta) &= \mathbb{E}_{x \sim \mathcal{D}} \Bigg[ \frac{1}{G} \sum_{i=1}^G \min \Big( \rho_i A_i, \\
    &\quad \quad \text{clip}\left( \rho_i, 1 - \varepsilon, 1 + \varepsilon \right) A_i \Big) \Bigg], \\
    \text{where } \rho_i &= \frac{\pi_\theta(z_i, y_i \mid x)}{\pi_{\theta_{\text{old}}}(z_i, y_i \mid x)}.
\end{split}
\end{equation}

The advantage $A_i$ is computed as:
\begin{equation}
\begin{gathered}
A_i = \frac{r_i - \mu_r}{\sigma_r + \epsilon_{\text{norm}}}, \\
\mu_r = \frac{1}{G}\sum_{j=1}^{G} r_j, \quad \sigma_r = \sqrt{\frac{1}{G}\sum_{j=1}^{G}(r_j - \mu_r)^2}.
\end{gathered}
\end{equation}

\subsection{Group Pattern Selection Optimization (GPSO)}

We now present our proposed method, Group Pattern Selection Optimization (GPSO), which extends RLVR with the ability to explore and learn the most effective reasoning patterns for different problems. As shown in Figure \ref{fig:model}, the central idea is to leverage multiple candidate patterns appended to the prompt, evaluate their effectiveness using verifier-based rewards, and then selectively update the policy with the optimal pattern while preventing overfitting to pattern-related suffix tokens through attention masking.

\paragraph{Multi-Pattern Rollout.}
Given a prompt $x$, we introduce a set of $n$ reasoning patterns $\{p_1,\dots,p_n\}$. Each pattern serves as a suffix that encourages the model to follow a distinct reasoning trajectory. For each $p_j$, the policy $\pi_\theta$ samples $m$ responses:
\begin{equation}
\mathcal G_j = \{ y_{j,1}, y_{j,2}, \dots, y_{j,m} \} \sim \pi_\theta(\cdot \mid x \oplus p_j),
\end{equation}
where $\oplus$ denotes prompt concatenation. Each response $y_{j,k}$ receives a verifier reward $r_{j,k} = Verifier(y_{j,k}, y^*)$, $y^*$ is the golden answer.

\paragraph{Pattern Selection Rule.}
To determine the most effective reasoning strategy, we compute the empirical accuracy of each pattern:
\begin{equation}
Acc(p_j) = \frac{1}{m}\sum_{k=1}^m \mathbf{1}[r_{j,k}=1],
\end{equation}
and select the optimal pattern
\begin{equation}
p^* = \arg\max_{p_j} Acc(p_j).
\end{equation}
When multiple patterns achieve the same accuracy, we select the one producing the shortest valid reasoning trace $\ell(y_{j,k})$, favoring concise solutions. The responses guided by the selected pattern $p^*$ are then used to perform the subsequent policy update.

\paragraph{Attention Masking for Pattern Suffix.}
While suffixes $p_j$ guide exploration, we prevent the model from overfitting by masking out their contribution during gradient updates. Concretely, let $M \in \{0,1\}^{B \times (L_{\text{prompt}}+L_{\text{resp}})}$ be the attention mask, where $B$ is the batch size, $L_{\text{prompt}}$ is the maximum prompt length, and $L_{\text{resp}}$ is the maximum response length. For a given sequence, $M_{i,t}=0$ indicates that token $t$ in instance $i$ is masked out, and $M_{i,t}=1$ otherwise. In particular, for tokens corresponding to the appended pattern suffix, we enforce
\begin{equation}
M_{i,t} = 0, \quad \forall t \in \text{Idx}(p_j),
\end{equation}
where $\text{Idx}(p_j)$ denotes the index set of token positions occupied by suffix $p_j$. This ensures that suffix tokens cannot influence the contextual representation of other tokens. Thus, patterns act as exploration scaffolds but do not directly leak into the learned policy.

\paragraph{Training Objective.}
Once $p^*$ is identified, we restrict optimization to its sampled group $\mathcal G_{p^*}$. Let $\hat A_{p^*,k}$ denote the group-normalized advantage, computed as in GRPO but masked such that gradient flow ignores suffix positions. Formally, the GPSO objective is:

\begingroup


\begin{equation}
\begin{aligned}
    \mathcal{L}_{\text{GPSO}}(\theta) &= \mathbb{E}_{x \sim \mathcal{D}} \Bigg[
    \frac{1}{\mathcal{G}_{p^*}} \sum_{k=1}^{\mathcal{G}_{p^*}} \min \Big(
        \rho_k \hat{A}_{p^*,k}, \\
    &\quad \quad \operatorname{clip}\left( \rho_k, 1 - \varepsilon, 1 + \varepsilon \right) \hat{A}_{p^*,k}
    \Big) \Bigg], \\
    \text{where } \rho_k &= \frac{\pi_{\theta}(z_k, y_k \mid x \oplus p^*)}{\pi_{\theta_{\text{old}}}(z_k, y_k \mid x \oplus p^*)}.
\end{aligned}
\end{equation}

\endgroup

Here, $\hat A_{p^*,k}$ is computed by normalizing rewards within $\mathcal G_{p^*}$, and gradients are masked to exclude suffix tokens. In this way, GPSO leverages pattern-based exploration to discover effective reasoning trajectories while maintaining a clean separation between exploration scaffolds and the policy itself.

\section{Experimental Methodology}
\label{sec:experiment}
In this section, we describe the datasets, evaluation metrics, baselines, and implementation details.

\textbf{Dataset. }
For the training set, we use DAPO-Math-17K dataset \citep{yu2025dapoopensourcellmreinforcement}, which is a curated collection of approximately 17,000 competition-level math problems. For testing, we evaluate the effectiveness of GPSO on AIME2024 \citep{numina_math_datasets}, AIME2025 \citep{ye2025limoreasoning}, MATH-500 \citep{hendrycksmath2021}, and GPQA datasets \citep{rein2023gpqagraduatelevelgoogleproofqa}.

\textbf{Evaluation Metrics. }
We follow previous work \citep{chen2021evaluatinglargelanguagemodels,li2024mmcode,wang2024intervenor,yang2024enhancing,luo2023wizardcoder} and we use \texttt{Pass@k} \citep{chen2021evaluatinglargelanguagemodels} to evaluate the effectiveness of different models. In this work, we set $k=1$. The \texttt{Pass@1} accuracy is averaged over 4 samples per problem on all benchmarks.

\begin{table*}[t]
\centering

\vspace{0.4mm}

\small
\setlength{\tabcolsep}{2.4pt}      
\renewcommand{\arraystretch}{0.86} 

\resizebox{\linewidth}{!}{
\begin{tabular}{>{\bfseries}lccccc}
\toprule
Model & \textbf{AIME2024} & \textbf{AIME2025} & \textbf{MATH500} & \textbf{GPQA} & \textbf{Avg.} \\
\midrule
DeepSeek-R1-Distill-Qwen-1.5B        & 30.0 & 20.0 & 84.7 & 33.8 & 42.1 \\
DeepScaleR-1.5B-Preview              & 40.2 & 28.5 & 87.8 & 32.3 & 47.2 \\
Light-R1-7B-DS                       & 57.7 & 46.4 & 91.1 & 47.2 & 60.6 \\
AReal-boba-RL-7B                     & 62.7 & 49.4 & 93.8 & 48.0 & 63.5 \\
DeepSeek-R1-Distill-Qwen-14B         & 70.4 & 50.0 & 92.4 & 59.5 & 68.1 \\
\midrule
Nemotron-Research-Reasoning-Qwen-1.5B & 53.3 & 35.8 & 92.1 & 40.5 & 55.4 \\
\hspace{0.6em}+ GPSO                  & \textbf{58.3} & \textbf{37.5} & \textbf{93.1} & \textbf{43.2} & \textbf{58.0} \\
\midrule
DeepSeek-R1-Distill-Qwen-7B           & 48.3 & 33.3 & 93.2 & 47.6 & 55.6 \\
\hspace{0.6em}+ GPSO                  & \textbf{53.3} & \textbf{40.0} & \textbf{93.5} & \textbf{47.9} & \textbf{58.7} \\
\midrule
DeepSeek-R1-Distill-Llama-8B          & 44.2 & 27.5 & 88.1 & 46.0 & 51.4 \\
\hspace{0.6em}+ GPSO                  & \textbf{49.2} & \textbf{29.2} & \textbf{90.2} & \textbf{50.0} & \textbf{54.6} \\
\midrule
Qwen3-8B (Thinking)                   & 76.7 & 67.5 & 96.0 & 58.0 & 74.5 \\
\hspace{0.6em}+ GPSO                  & \textbf{77.5} & \textbf{68.3} & \textbf{96.1} & \textbf{59.2} & \textbf{75.3} \\
\bottomrule
\end{tabular}
}
\caption{Overall performance of Group Pattern Selection Optimization (GPSO).}
\label{tab:overall-performance}
\end{table*}

\textbf{Baselines. }
We compare GPSO with several LRMs, such as
DeepSeek-R1-Distill-Qwen-1.5B/14B \citep{deepseekai2025deepseekr1incentivizingreasoningcapability},
DeepScaleR-1.5B-Preview \citep{deepscaler2025},
Light-R1-7B-DS \citep{wen2025lightr1curriculumsftdpo},
AReal-boba-RL-7B \citep{fu2025areallargescaleasynchronousreinforcement}. DeepScaleR-1.5B-Preview is further trained starting from DeepSeek-R1-Distill-Qwen-1.5B, while Light-R1-7B-DS and AReal-boba-RL-7B are further trained from DeepSeek-R1-Distill-Qwen-7B.

\textbf{Implementation Details. }
In our experiments, we apply GPSO to four LRMs: Nemotron-Research-Reasoning-Qwen-1.5B \citep{liu2025prorlprolongedreinforcementlearning}, DeepSeek-R1-Distill-Qwen-7B \citep{deepseekai2025deepseekr1incentivizingreasoningcapability}, DeepSeek-R1-Distill-Llama-8B \citep{deepseekai2025deepseekr1incentivizingreasoningcapability}, and Qwen3-8B (Thinking) \citep{qwen3technicalreport}. 
During training, we use Verl framework \citep{sheng2024hybridflow} and apply GRPO as the RL algorithm to implement GPSO. For hyperparameters, we set the batch size and mini-batch size to $64$, and for each problem, we rollout $8$ responses using four patterns: Direct Solution, Reflection and Verification, Exploration of Multiple Solutions, and Adaptive. For baselines, we rollout $32$ responses for each question to ensure fair comparison. The prompts we used are in Appendix~\ref{app:prompts}. The maximum lengths for prompts and responses are $1,024$ and $16,384$ tokens, respectively. The learning rate is set to $1e-6$, and we adopt the AdamW optimizer for the policy model.  During testing, we set the temperature to $0.6$. The maximum generation length is set to $32,768$ tokens for AIME 2024/2025 and $16,384$ tokens for MATH-500 and GPQA. All evaluations are conducted under the zero-shot setting.

\section{Evaluation Results}
In this section, we present the evaluation results for GPSO. Our evaluation includes a comprehensive analysis of the overall performance, ablation studies to assess the contribution of key components, and insights into how GPSO enhances reasoning performance across a variety of tasks.

\subsection{Overall Performance}
The overall performance of GPSO is shown in Table \ref{tab:overall-performance}. Across different model backbones, applying GPSO consistently improves performance. Nemotron-Research-Reasoning-Qwen-1.5B improves its average score from 55.4 to 58.0 (+2.6\%), while DeepSeek-R1-Distill-Qwen-7B increases from 55.6 to 58.7 (+3.1\%). Similarly, DeepSeek-R1-Distill-Llama-8B improves from 51.4 to 54.6 (+3.2\%). Notably, Qwen3-8B (Thinking) further benefits from GPSO, achieving the best overall average of 75.3. These results indicate that GPSO is model-agnostic and provides stable gains. On individual benchmarks, the improvements brought by GPSO mainly come from challenging reasoning tasks such as AIME 2024 and AIME 2025. Across all four models, GPSO yields an average gain of 4.0~points on AIME2024 and 2.7~points on AIME2025. Moreover, although GPSO is trained solely on mathematical data, it demonstrates strong generalization across domains, achieving an average improvement of 2.1~points on GPQA. These results confirm that GPSO offers a plug-and-play enhancement to existing RLVR training pipelines, with consistent gains across both weak and strong LLMs.

\begin{table*}[t]
\centering

\vspace{1mm}

\resizebox{\linewidth}{!}{
\begin{tabular}{lcccccccc}
\toprule
\textbf{Model}                                     & \textbf{MPR} & \textbf{OPS} & \textbf{Mask} & \textbf{KL}& \textbf{AIME2024} & \textbf{AIME2025} & \textbf{GPQA} & \textbf{Avg} \\ \hline
\textbf{Nemotron-Qwen-1.5B}     & -          & -              & -        & -      & 53.3           & 35.8                 & 40.5             & 43.2            \\
\textbf{Nemotron-Qwen-1.5B-GPSO} & \correct        & \correct            & \correct      & \wrong                 & \textbf{58.3}                 & \textbf{37.5}             & \textbf{43.2}      & \textbf{46.3}      \\
\textbf{\hspace{1em}w/ KL}                 & \correct        & \correct            & \correct        & \correct                 & 54.2          & 36.7       & 40.8             & 43.9            \\
\textbf{\hspace{1em}w/o Multi-Pattern Rollout}                 & \wrong          & \wrong              & \wrong        & \correct                 & 49.2          & 33.3       & 40.5             & 41.0            \\
\textbf{\hspace{1em}w/o Optimal Pattern Selection}             & \correct        & \wrong              & \correct      & \correct                 & 53.3          & 35.8       & 40.0             & 43.1            \\
\textbf{\hspace{1em}w/o Mask Pattern Tokens}                   & \correct        & \correct            & \wrong        & \correct                 & 56.7       & 33.3          & 40.3             & 43.4            \\ \hline
\textbf{DeepSeek-R1-Distill-Qwen-7B}               & -          & -              & -        & -                 & 48.3                 & 33.3    & 47.6         & 43.2            \\
\textbf{DeepSeek-R1-Distill-Qwen-7B-GPSO}           & \correct        & \correct            & \correct      & \wrong               & \textbf{53.3}                 & \textbf{40.0}     & 47.9        & \textbf{47.1}            \\
\textbf{\hspace{1em}w/ KL}                 & \correct        & \correct            & \correct        & \correct                 & 52.5          & 36.7       & 46.3             & 45.2            \\
\textbf{\hspace{1em}w/o Multi-Pattern Rollout}                 & \wrong          & \wrong              & \wrong        & \correct                 & 50.8        & 35.8         & 47.7             & 44.8            \\
\textbf{\hspace{1em}w/o Optimal Pattern Selection}             & \correct        & \wrong              & \correct      & \correct                 & 50.8       & 38.3          & \textbf{50.3}             & 46.5            \\
\textbf{\hspace{1em}w/o Mask Pattern Tokens}                   & \correct        & \correct            & \wrong        & \correct                 & 51.7      & 38.3           & 45.7             & 45.2            \\ \hline
\end{tabular}
}
\caption{Ablation Studies. We evaluate the impact of removing each component in GPSO: Multi-Pattern Rollout (MPR), Optimal Pattern Selection (OPS), Masking Pattern Tokens (Mask), and the KL penalty (KL). \correct \ indicates the component is enabled, while \wrong \   indicates it is disabled.}
\label{tab:abla}
\end{table*}
\begin{figure}[t] \centering
    \includegraphics[width=0.5\textwidth]{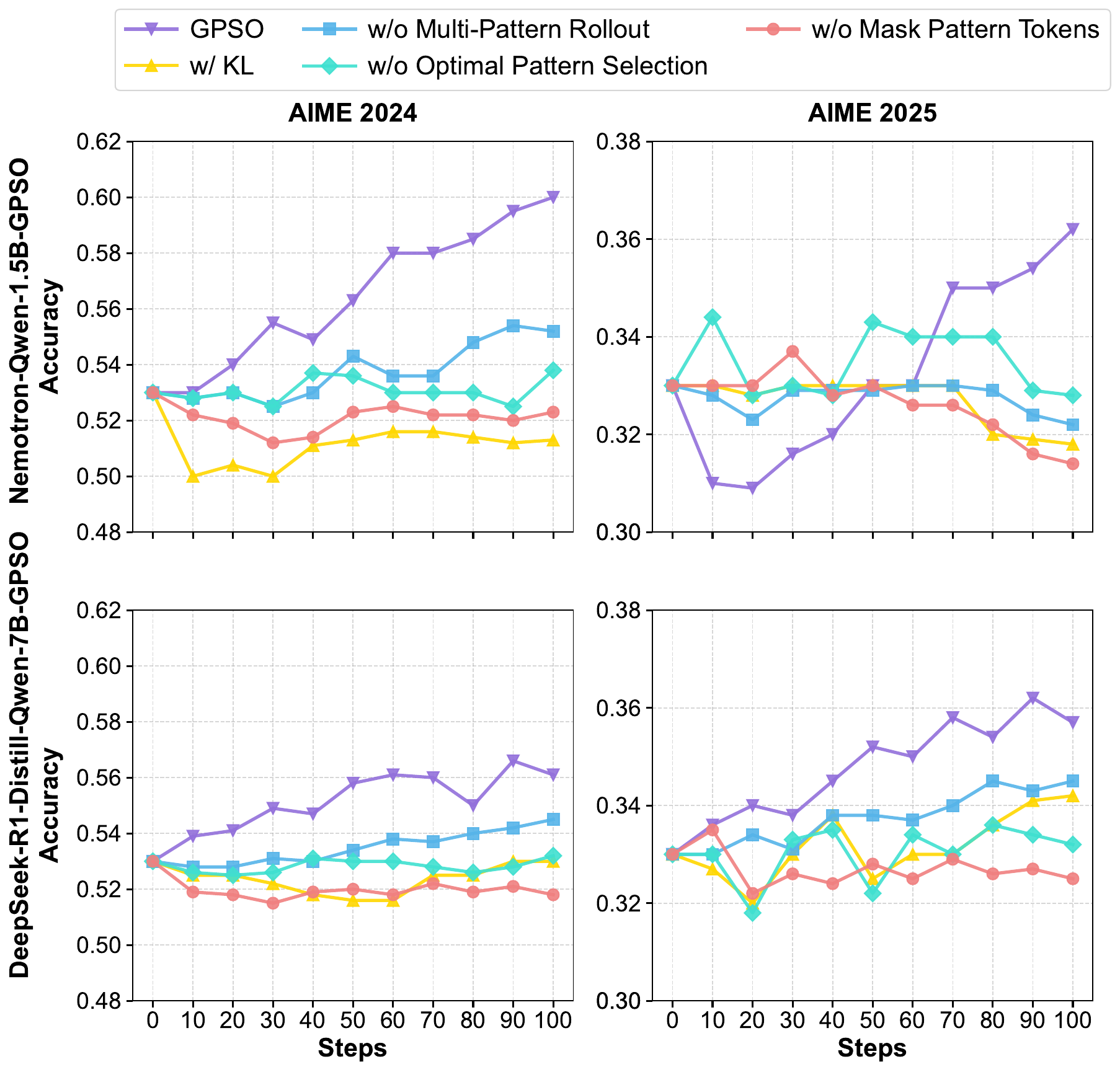}
    \caption{Training Accuracy Curves On AIME2024 and AIME2025.} \label{fig:abla}
\end{figure}
\subsection{Ablation Studies}
To further investigate the individual contributions of the key components in GPSO, we conduct a series of ablation experiments. As shown in Table \ref{tab:abla} and Figure \ref{fig:abla}, we evaluate the model under several settings: (1) removing the KL penalty, (2) excluding the Multi-Pattern Rollout mechanism, (3) disabling the Optimal Pattern Selection, and (4) not masking the Pattern Tokens. From the results and training accuracy curves on AIME2024 and AIME2025, we observe that removing any of these components leads to a noticeable performance degradation.

We summarize our key observations as follows. First, the \textbf{KL penalty} consistently hurts performance across both models and datasets. As shown in Figure \ref{fig:abla}, the training curves with KL remain persistently below the GPSO training, indicating that the KL regularization constrains the policy from adequately exploring the solution space and learning effective task-specific behaviors. This is also reflected in Table \ref{tab:abla}, where models trained with the KL penalty systematically underperform their counterparts trained without it. Specifically, removing the KL penalty yields an average improvement of 2.4 points for Nemtorn-Qwen-1.5B-GPSO and 1.9 points for DeepSeek-R1-Distill-Qwen-7B-GPSO, demonstrating that the KL regularization imposes overly strong constraints on policy updates and hinders the models' ability to adapt meaningfully during the optimization process. Second, excluding the \textbf{Multi-Pattern Rollout mechanism} leads to the most significant performance degradation. Without rollout, the model struggles to explore diverse reasoning paths and quickly plateaus. The average performance drops by 5.3 points (compared with Nemtorn-Qwen-1.5B-GPSO) and 2.3 points (compared with DeepSeek-R1-Distill-Qwen-7B-GPSO), highlighting the critical role of this component in guiding exploration. Third, turning off \textbf{Optimal Pattern Selection} results in a moderate but consistent decrease. Although the Multi-Pattern Rollout still runs, the lack of selection prevents the model from reinforcing high-quality patterns, leading to noisier supervision. This is most noticeable on AIME2025, where accuracy deteriorates by 1.7 points on both models. Lastly, we observe that \textbf{Masking Pattern Tokens} also plays a subtle but meaningful role. Without this masking, the model has access to hard-coded pattern identifiers, which may introduce undesirable shortcuts during learning. Both Table \ref{tab:abla} and Figure \ref{fig:abla} show that disabling masking results in slower convergence and slightly worse final performance, suggesting that overfitting to pattern identity is more detrimental in harder, unfamiliar tasks.

\begin{table*}[t]
\centering

\vspace{1mm}

\resizebox{\linewidth}{!}{
\begin{tabular}{llcccc}
\toprule
\textbf{Model}                                            & \textbf{Pattern}            & \textbf{AIME2024} & \textbf{AIME2025} & \multicolumn{1}{l}{\textbf{MATH500}} & \multicolumn{1}{l}{\textbf{GPQA}} \\ \hline
\multirow{4}{*}{\textbf{Nemotron-Qwen-1.5B}}              & -                           & 53.3            & 33.3            & 92.1                               & 40.5                            \\
                                                          & Direct Solution             & 41.7            & 27.5            & 90.1                               & 39.3                            \\
                                                          & Reflection and Verification & 47.5            & 30.8            & \textbf{92.2}                      & \textbf{41.3}                   \\
                                                          & Explore Multiple Solutions  & \textbf{55.0}   & \textbf{34.2}   & 91.7                               & 38.6                            \\ \hline
\multirow{4}{*}{\textbf{Nemotron-Qwen-1.5B-GPSO}}          & -                           & \textbf{58.3}   & \textbf{37.5}   & \textbf{93.1}                      & \textbf{43.2}                   \\
                                                          & Direct Solution             & 49.2            & 33.3            & 91.1                               & 40.5                            \\
                                                          & Reflection and Verification & 53.3            & 35.8            & 91.6                               & 40.0                            \\
                                                          & Explore Multiple Solutions  & 56.7            & 33.3            & 91.3                               & 40.3                            \\ \hline 
\multirow{4}{*}{\textbf{DeepSeek-R1-Distill-Qwen-7B}}     & -                           & 48.3            & 33.3            & \textbf{93.2}                      & \textbf{47.6}                   \\
                                                          & Direct Solution             & 43.3            & 26.7            & 88.1                               & 44.8                            \\
                                                          & Reflection and Verification & \textbf{50.0}   & 33.3            & 92.6                               & \textbf{47.6}                   \\
                                                          & Explore Multiple Solutions  & 48.3            & \textbf{36.7}   & 90.3                               & 47.5                            \\ \hline 
\multirow{4}{*}{\textbf{DeepSeek-R1-Distill-Qwen-7B-GPSO}} & -                           & \textbf{53.3}   & \textbf{40.0}   & \textbf{93.5}                      & \textbf{47.9}                   \\
                                                          & Direct Solution             & 45.0            & 36.7            & 90.2                               & 46.5                            \\
                                                          & Reflection and Verification & 49.2            & 38.3            & 92.8                               & 47.4                            \\
                                                          & Explore Multiple Solutions  & 51.7            & 39.2            & 92.2                               & 47.0                            \\ \hline
\end{tabular}
}
\caption{Effectiveness of Reasoning Pattern Selection with GPSO}
\label{tab:learning}
\end{table*}
\subsection{GPSO learns to pick the right pattern for reasoning}

As shown in Table~\ref{tab:learning}, GPSO enables both models to dynamically apply the most suitable reasoning pattern per instance, outperforming all fixed-pattern baselines. Without GPSO, no single reasoning mode consistently dominates across benchmarks. For example, Nemotron-Qwen-1.5B performs best with Explore Multiple Solutions on AIME2024 and AIME2025, but achieves higher scores with Reflection and Verification on MATH500 and GPQA. DeepSeek-R1-Distill-Qwen-7B shows similar variability. In contrast, GPSO-trained models achieve the highest scores across all benchmarks using the default decoding strategy—surpassing even the best fixed-pattern results. This demonstrates that GPSO can effectively learn to adaptively combine reasoning strategies based on the problem type, leading to more generalizable and robust performance.

\begin{figure*}[t] \centering
    \includegraphics[width=1\textwidth]{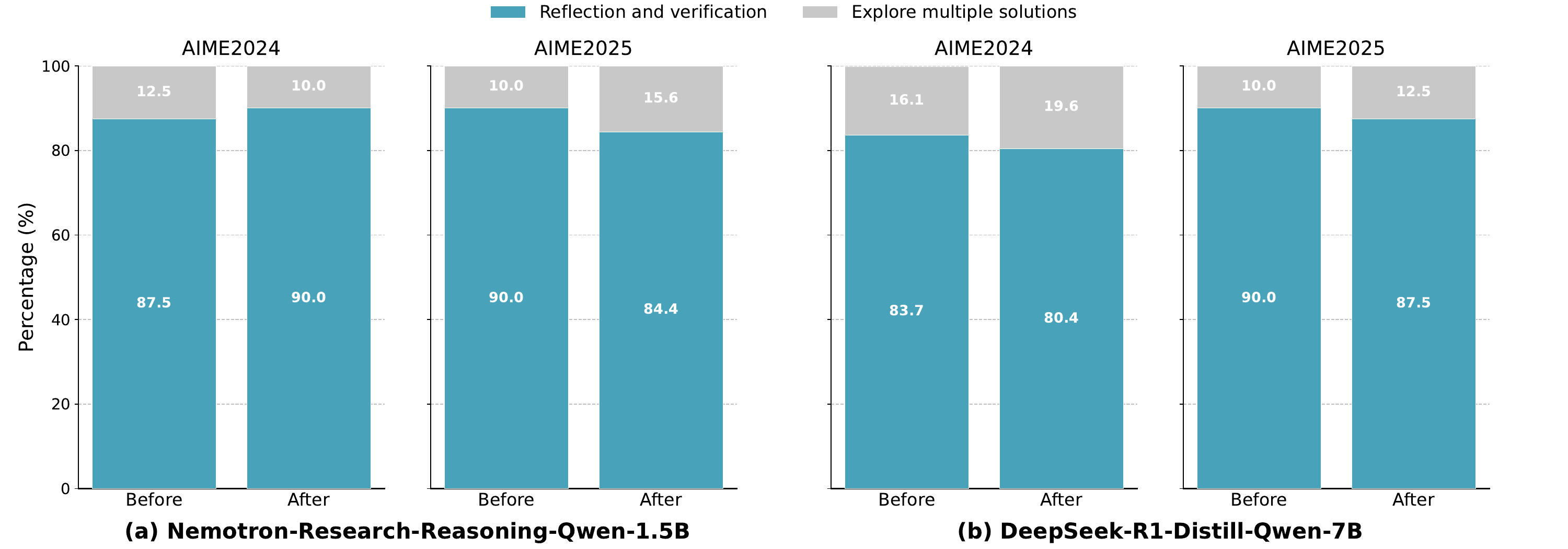}
    \caption{Pattern Usage Distribution Before and After GPSO Training} \label{fig:pattern}
\end{figure*}
\subsection{Distribution of reasoning patterns before and after GPSO training.}
As shown in Figure~\ref{fig:pattern}, we analyze the distribution of reasoning patterns selected on the AIME2024 and AIME2025 datasets, both before and after applying GPSO. The results confirm that GPSO enables models to learn an adaptive, problem-dependent policy rather than converging to a single fixed strategy.

For Nemotron-Research-Reasoning-Qwen-1.5B (Figure~\ref{fig:pattern}(a)), we observe a clear task-specific adjustment. On AIME2024, the model further strengthens its preference for the Reflection and Verification pattern, increasing its usage from 87.5\% to 90.0\%. In contrast, on the more challenging AIME2025, the model shifts towards Explore Multiple Solutions, increasing its usage from 10.0\% to 15.6\%. This indicates that GPSO guides the model to adopt more exploratory strategies when the problem requires it. A similar trend is observed with DeepSeek-R1-Distill-Qwen-7B (Figure~\ref{fig:pattern}(b)). On AIME2024, the share of Explore Multiple Solutions rises from 16.1\% to 19.6\%, and on AIME2025, from 10.0\% to 12.5\%. These shifts further highlight GPSO’s ability to learn a meta-policy that adjusts the invocation probabilities of different reasoning strategies based on task characteristics.

\section{Conclusion}
In this work, we propose GPSO, a novel training paradigm that enables language models to select optimal reasoning patterns per instance dynamically. By combining multi-pattern exploration, verifier-guided supervision, and gradient-masked updates, GPSO teaches the model to internalize reasoning strategies without relying on explicit prompts. Experiments on multiple benchmarks demonstrate that GPSO consistently improves performance across models and tasks, particularly on challenging datasets that require reasoning. Our results highlight the effectiveness of adaptive pattern selection in enhancing both accuracy and generalization of LLM reasoning.
\section*{Limitations}
While GPSO effectively optimizes reasoning patterns, it acknowledges several limitations. First, the training process involves multi-pattern rollouts for each problem, which increases the computational cost compared to standard single-path reinforcement learning. Second, the set of candidate reasoning patterns is predefined based on empirical observations; exploring mechanisms for automatically discovering or evolving new patterns remains a promising direction for future research.

\section*{Ethics Statement}
We propose GPSO to improve LLM reasoning efficiency. Our experiments rely exclusively on publicly available academic benchmarks (e.g., AIME, GPQA) containing no personally identifiable information. We acknowledge that advanced reasoning capabilities carry potential dual-use risks; however, our method utilizes verifiable rewards based on objective mathematical truths, minimizing the risk of hallucinating harmful content during training. We are committed to open science and will release our code and artifacts to ensure reproducibility.

\section*{LLM Use}
We used LLMs for polishing the text and improving the readability of the manuscript.

\bibliography{custom}

\appendix
\newpage

\appendix

\section{Distribution of Reasoning Patterns Across Domains}
\label{app:distribution}

To support our analysis, we sample 1,000 reasoning trajectories from seven LRMs for the mathematics and science domains. Each response is annotated into one of five high-level reasoning categories: Direct Solution, Explore Multiple Solutions, Reflection and Verification, Analogy, and Reverse Thinking.

\begin{figure*}[t]
    \centering
    \includegraphics[width=1\linewidth]{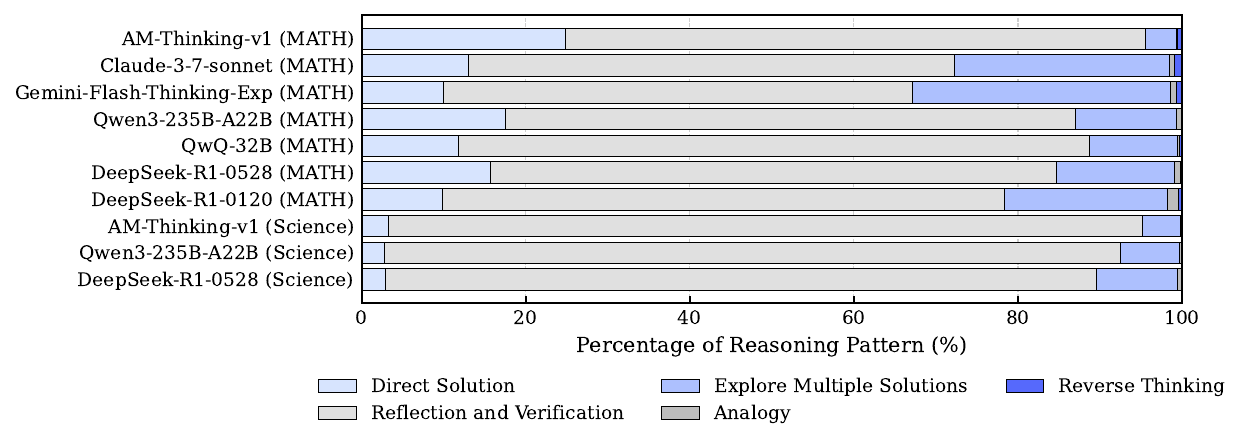}
    \caption{Distribution of reasoning patterns used by various LLMs on MATH and Science tasks}
    \label{fig:dis}
\end{figure*}

Figure \ref{fig:dis} provides a detailed breakdown of the percentage distribution of reasoning patterns exhibited by each model. We observe clear trends—such as the dominance of Reflection and Verification in most models, particularly in the science domain, and the relatively lower adoption of Explore Multiple Solutions or Reverse Thinking. These results underscore the tendency of LLMs to default to a small subset of reasoning strategies, despite their architectural capacity for diverse reasoning.

\section{Prompts}
\label{app:prompts}

\subsection{Full Prompts for Pattern Analysis}
\begin{tcolorbox}[promptbox, title={Full Prompt for Response Classification}]
I will provide you with a problem and its solution. Please analyze the strategy used in this solution.

When solving problems, people may use the following strategies:
\begin{itemize}
  \item \textbf{Direct solution}: Solve step by step without reflection or verification, then give the answer.
  \item \textbf{Reflection and verification}: Reflect and verify; if errors are found, correct and provide the final answer.
  \item \textbf{Explore multiple solutions}: Try multiple approaches and select/synthesize to give the final answer.
  \item \textbf{Others}: Not among the above.
\end{itemize}

Please select the \emph{single} strategy that is most appropriate.

\medskip
\textbf{Problem:} \texttt{\{question\}}\\
\textbf{Solution:} \texttt{\{solution\}}

\medskip
In your response, first conduct the analysis and then output:
\texttt{<strategy> ... </strategy>} (tags contain only the strategy name).
\end{tcolorbox}

\subsection{Prompt Examples for Pattern Reasoning}

\begin{tcolorbox}[promptbox, title={No Constraint}]
\PromptToyProblem
\end{tcolorbox}

\begin{tcolorbox}[promptbox, title={Direct Solutions}]
\PromptToyProblem
\tcblower
\begin{tcolorbox}[promptinst]
\textbf{Instruction.} Provide the final answer concisely. Do not self-reflect or self-criticize.
\end{tcolorbox}
\end{tcolorbox}

\begin{tcolorbox}[promptbox, title={Explore Multiple Solutions}]
\PromptToyProblem
\tcblower
\begin{tcolorbox}[promptinst]
\textbf{Instruction.} Explore multiple solutions, then provide the final answer based on them.
\end{tcolorbox}
\end{tcolorbox}

\begin{tcolorbox}[promptbox, title={Reflection and Verication}]
\PromptToyProblem
\tcblower
\begin{tcolorbox}[promptinst]
\textbf{Instruction.} Reflect on your reasoning and verify correctness before the final answer.
If you find errors, correct them and then provide the final answer.
\end{tcolorbox}
\end{tcolorbox}

\end{document}